\documentclass{jingdong}
\usepackage{amsmath,amssymb}
\usepackage{array}
\usepackage{colortbl}
\usepackage[shortlabels,inline]{enumitem}
\usepackage{algorithm}
\usepackage{algpseudocode}
\newcommand{\resourcecodeicon}{%
  \tikz[baseline=-0.12em,x=0.85em,y=0.85em,line width=0.6pt,line cap=round,line join=round]{%
    \draw (0.25,0.15) -- (0,0.4) -- (0.25,0.65);
    \draw (0.75,0.15) -- (1,0.4) -- (0.75,0.65);
    \draw (0.4,0) -- (0.6,0.8);
  }}
\newcommand{\resourcemodelicon}{%
  \tikz[baseline=-0.12em,x=0.85em,y=0.85em,line width=0.6pt,line join=round]{%
    \draw (0.45,0) -- (0.9,0.25) -- (0.9,0.75) -- (0.45,1) -- (0,0.75) -- (0,0.25) -- cycle;
    \draw (0,0.75) -- (0.45,0.5) -- (0.9,0.75);
    \draw (0.45,0.5) -- (0.45,0);
  }}
\newcommand{\resourcedataicon}{%
  \tikz[baseline=-0.12em,x=0.85em,y=0.85em,line width=0.6pt,line join=round]{%
    \draw (0.45,0.75) ellipse [x radius=0.45,y radius=0.15];
    \draw (0,0.75) -- (0,0.15) arc [start angle=180,end angle=360,x radius=0.45,y radius=0.15] -- (0.9,0.75);
    \draw (0,0.45) arc [start angle=180,end angle=360,x radius=0.45,y radius=0.15];
  }}
\crefname{figure}{Fig.}{Figs.}
\Crefname{figure}{Fig.}{Figs.}
\crefname{table}{Table}{Tables}
\Crefname{table}{Table}{Tables}

\title{Turns Matter: Advancing Streaming Video RL via Turn-Level Credit}

\author[1,2,3,*]{Shuai Dong}
\author[1,3,5,*]{Kanghui Tian}
\author[1,*]{Linghan Chen}
\author[3,\ddagger]{Wenqi Shao}
\author[3,4]{Hongrui Wu}
\author[2,3]{Haowen Hou}
\author[3]{Yongfu Zhu}
\author[3]{Xuehui Wang}
\author[3,7]{Ziyue Wang}
\author[3,7]{Wenpu Liu}
\author[3,7]{Weichu Xie}
\author[3]{Yuqi Xu}
\author[3]{Congcong Wang}
\author[3,9]{Kaiwen Tuo}
\author[3,8]{Caoyuan Ma}
\author[3,10]{Taiqiang Wu}
\author[2,3]{Ji Zhao}
\author[3,5]{Wenzheng Chang}
\author[1,2]{Dianyi Wang}
\author[6,\dagger]{Siyuan Wang}
\author[1,2,\dagger]{Zhongyu Wei}
\author[2,5]{Yu Qiao}
\author[3]{Jiaqi Wang}

\affiliation[1]{Fudan University}
\affiliation[2]{Shanghai Innovation Institute}
\affiliation[3]{JD.com}
\affiliation[4]{Stanford University}
\affiliation[5]{Shanghai Artificial Intelligence Laboratory}
\affiliation[6]{The Chinese University of Hong Kong}
\affiliation[7]{Peking University}
\affiliation[8]{The University of Tokyo}
\affiliation[9]{The Hong Kong University of Science and Technology}
\affiliation[10]{The University of Hong Kong}

\contribution[*]{Equal contribution}
\contribution[\ddagger]{Project leader}
\contribution[\dagger]{Corresponding authors.}

\resources{%
  {\fontsize{12}{14}\selectfont\sffamily\bfseries
  \mbox{\resourcecodeicon\hspace{0.4em}\href{https://github.com/XD111ds/turn-wise-rl}{Code}}%
  \hspace{2em}%
  \mbox{\resourcemodelicon\hspace{0.4em}\href{https://huggingface.co/shuai22/turns-matter-rl-ckpt}{Model}}%
  \hspace{2em}%
  \mbox{\resourcedataicon\hspace{0.4em}\href{https://huggingface.co/datasets/shuai22/mmduet2-training-data}{Dataset}}}
}

\hypersetup{
    pdftitle={Turns Matter: Advancing Streaming Video RL via Turn-Level Credit},
    pdfauthor={Shuai Dong, Kanghui Tian, Linghan Chen, Wenqi Shao, Hongrui Wu, Haowen Hou, Yongfu Zhu, Xuehui Wang, Ziyue Wang, Wenpu Liu, Weichu Xie, Yuqi Xu, Congcong Wang, Kaiwen Tuo, Caoyuan Ma, Taiqiang Wu, Ji Zhao, Wenzheng Chang, Siyuan Wang, Zhongyu Wei, Yu Qiao, Jiaqi Wang},
    pdfsubject={Preprint}
}

\abstract{\begingroup\looseness=-1Streaming video understanding requires models to repeatedly decide whether to respond or continue observing. Recently, Group Relative Policy Optimization (GRPO) has been applied to this setting, with existing approaches emphasizing reward design while retaining a single trajectory-level advantage across turns. Such shared credit can reinforce redundant replies after earlier successes and discourage useful follow-ups after earlier failures. We propose TurnWise, a turn-level GRPO framework that aligns learning with individual streaming decisions. TurnWise addresses the mismatch between trajectory-level feedback and turn-level interaction, promoting informative responses and reducing redundancy. Specifically, turn-level credit assignment uses silence-completed prefix scoring to construct group-relative advantages from each decision's remaining return, separating its learning signal from earlier rewards. Turn-mean policy optimization averages token losses within each turn and then across turns, giving response and waiting decisions equal weight regardless of output length. These complementary designs improve streaming policy learning without a learned critic or additional rollouts. Extensive experiments show our method outperforms prior RL approaches for streaming video understanding.
\par\endgroup}

\begin{document}
\maketitle
\begin{figure}[htbp]
\centering
\begingroup
\setlength{\fboxrule}{0.4pt}
\setlength{\fboxsep}{3pt}
\fbox{\includegraphics[width=\dimexpr.95\linewidth-2\fboxsep-2\fboxrule\relax,trim=0 35bp 0 0,clip]{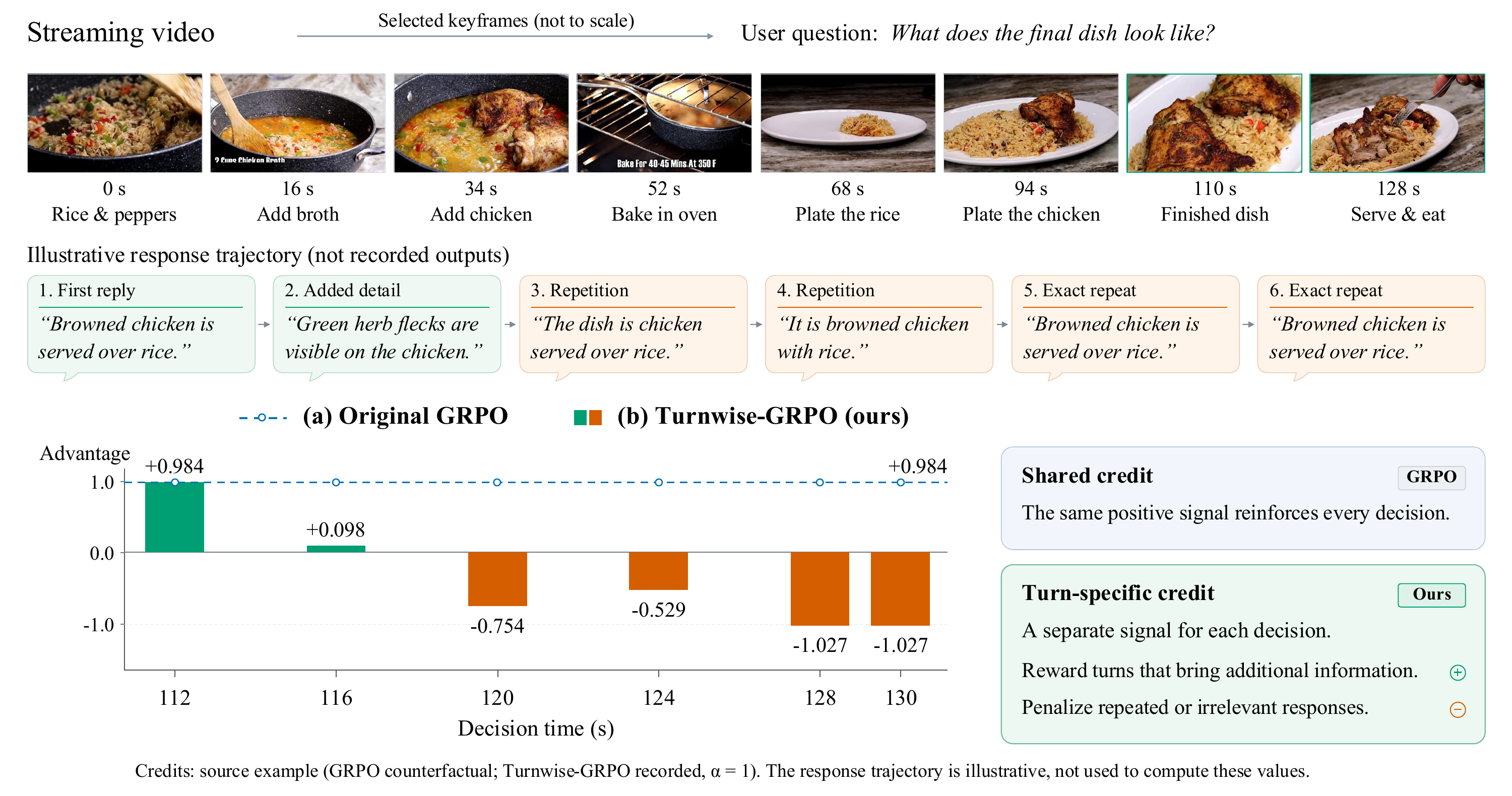}}
\par
\endgroup
\caption{\textbf{Shared advantages can reinforce redundant replies.} GRPO assigns one trajectory-level advantage to all turns, so later repetitions may inherit positive credit from earlier successes. TurnWise subtracts group-relative accrued reward before each decision, allowing later turns to receive credit according to their remaining return. The dialogue is illustrative. The advantage plot uses recorded TurnWise advantages and the corresponding GRPO counterfactual.}
\label{fig:teaser}%
\end{figure}

\section{Introduction}
\label{sec:intro}

Streaming video understanding requires models to interpret an evolving visual world and interact while observations are still arriving~\cite{chen2024videollmonline,qian2025dispider,zeng2026onestreamer}.
Unlike offline video understanding~\cite{zhang2024llavavideo}, where the complete video is available before answering, streaming interaction unfolds through repeated decisions under partial observations.
At each turn, a model must decide whether to respond or continue observing~\cite{wang2024mmduet,wang2025mmduet2}.
A useful assistant should respond when new evidence warrants an answer, provide follow-up information when needed, and remain silent when another reply would add little value.
The appropriate action therefore depends not only on the current visual evidence~\cite{dong2026interleaved,guo2026beyond,yu2026mact}, but also on what the model has already communicated.
Learning this sequence of response and waiting decisions requires feedback that reflects answer quality and interaction timing.

Reinforcement learning~\cite{sutton2018reinforcement,schulman2017ppo,shao2024deepseekmath,dong2026rl} provides a natural way to learn these decisions from interaction-level feedback, with recent approaches adopting Group Relative Policy Optimization (GRPO)~\cite{shao2024deepseekmath}.
MMDuet2~\cite{wang2025mmduet2} optimizes proactive interaction with PAUC-based rewards~\cite{wang2025proactivevideoqa}, while ThinkStream~\cite{liu2026thinkstream} combines incremental reasoning with rewards for correctness, format, and response timing.
StreamPro~\cite{li2026streampro} further incorporates both turn-level and trajectory-level rewards to assess individual responses and overall interaction quality.
Reward design and short-span training~\cite{wang2025mmduet2} improve interaction feedback, but the trajectory-level GRPO formulations considered here still broadcast one advantage across decision turns.
Turn-level scoring and turn-specific policy credit are therefore distinct design choices.

A good trajectory can contain bad turns, and a bad trajectory can contain good turns.
After a successful early response, a redundant reply may still inherit a positive trajectory advantage, even if the reward penalizes repetition (\cref{fig:teaser}).
Conversely, a useful follow-up after earlier failures may receive a negative advantage because the trajectory remains below the group average.
In both cases, rewards accrued before the current decision influence its learning signal, potentially obscuring the value of responding or waiting at that turn.
We therefore argue that a streaming decision's learning signal should emphasize the return from that turn onward, rather than rewards accumulated earlier.

Guided by this principle, we propose TurnWise, a turn-level GRPO framework that aligns advantage estimation and loss aggregation with streaming decisions.
Building on the return-to-go principle~\cite{sutton2018reinforcement}, TurnWise exploits the explicit waiting action to score each trajectory prefix with all remaining turns set to silence.
Subtracting this prefix reference from the full trajectory reward yields a remaining return, from which we construct group-relative, turn-specific advantages.
Silence-completed prefix scoring converts trajectory feedback into turn-specific remaining-return advantages.
The computation reuses existing rollout evaluations, requiring neither a learned critic nor additional rollouts.

TurnWise couples turn-specific credit with equal weighting of response and waiting decisions.
Under token-mean aggregation, longer responses still receive larger aggregation weights than short waiting actions.
We therefore introduce turn-mean loss aggregation, averaging token losses within each turn before averaging across turns within each trajectory.
Turn-mean aggregation balances sampled response and waiting decisions directly in the RL objective, making each turn's weight independent of its output length.
Together, the two designs make the decision turn a common unit for advantage estimation and loss aggregation, addressing both how a turn receives credit and how its feedback is weighted during policy optimization.

In summary, our contributions are as follows:
\begin{itemize}
    \item We identify a granularity mismatch in streaming video RL. Shared trajectory advantages can obscure the value of later decisions, while token-mean loss aggregation makes turn weights depend on response length.
    \item We introduce TurnWise, combining silence-completed prefix scoring for turn-specific remaining-return advantages with turn-mean loss aggregation, without a learned critic or additional rollouts.
    \item We achieve the best aggregate scores on ProactiveVQA~\cite{wang2025proactivevideoqa}, OVO-Timing~\cite{niu2025ovobench}, and StreamingBench~\cite{lin2026streamingbench} with both Qwen backbones, and show that turn-mean aggregation improves response timing while reducing repetition relative to credit-only optimization.
\end{itemize}

\section{Related Work}
\label{sec:related}

\subsection{Streaming Video Interaction}
Streaming video interaction extends offline video understanding~\cite{zhang2024llavavideo} by requiring models to process incoming observations and decide when to communicate.
Unlike answering after viewing an entire video, these models must operate on observed prefixes and accommodate evolving user queries.
VideoLLM-online~\cite{chen2024videollmonline} introduces online video-language modeling, while MMDuet~\cite{wang2024mmduet} broadens time-sensitive comprehension through interleaved video-text dialogue.
Dispider~\cite{qian2025dispider} disentangles perception, decision, and reaction, and StreamBridge~\cite{wang2025streambridge} equips offline models with a lightweight proactive activation module.
TimeChat-Online~\cite{yao2025timechatonline} reduces visual redundancy, while Flash-VStream~\cite{zhang2025flashvstream} maintains compact memory for real-time understanding of long video streams.
Streamo~\cite{xia2026streamo} further supports diverse streaming tasks through large-scale instruction tuning, including narration, event captioning, and time-sensitive question answering.
Benchmarks including ProactiveVideoQA~\cite{wang2025proactivevideoqa}, OVO-Bench~\cite{niu2025ovobench}, and StreamingBench~\cite{lin2026streamingbench} cover real-time perception, historical understanding, and proactive response timing.
These studies establish the setting for learning response and waiting decisions, while our work focuses on how RL feedback is assigned and weighted across those decisions.

\subsection{Reinforcement Learning for Video Understanding}
PPO~\cite{schulman2017ppo} and GRPO~\cite{shao2024deepseekmath} provide widely used foundations for language-model policy optimization.
Their application to video models has progressed from preference alignment to temporal reasoning and long-video training.
VLM-RLAIF~\cite{ahn2024vlmrlaif} uses AI-generated feedback for video instruction tuning.
Video-R1~\cite{feng2025videor1} encourages temporal reasoning through T-GRPO, VideoChat-R1~\cite{li2025videochatr1} strengthens spatio-temporal perception through reinforcement fine-tuning, and LongVILA-R1~\cite{chen2025scalingrl} scales RL training to long video inputs.
Policy distillation can improve initialization for subsequent RL~\cite{dong2026rl,yu2026dopd}.
In streaming settings, MMDuet2~\cite{wang2025mmduet2} optimizes PAUC-based interaction rewards and uses short-span training to mitigate reward sparsity and temporal credit assignment.
ThinkStream~\cite{liu2026thinkstream} trains incremental reasoning with accuracy, format, and timing rewards, while StreamPro~\cite{li2026streampro} combines turn-level response scores with trajectory-level coherence rewards.
These approaches improve evaluation of sampled interactions, but their trajectory-level GRPO formulations derive a shared advantage from the aggregated reward.
TurnWise aligns streaming policy optimization with decision turns through group-relative remaining-return credit and turn-mean loss aggregation.

\section{Methodology}
\label{sec:method}
\begin{figure}[tbp]
    \centering
    \includegraphics[width=\textwidth]{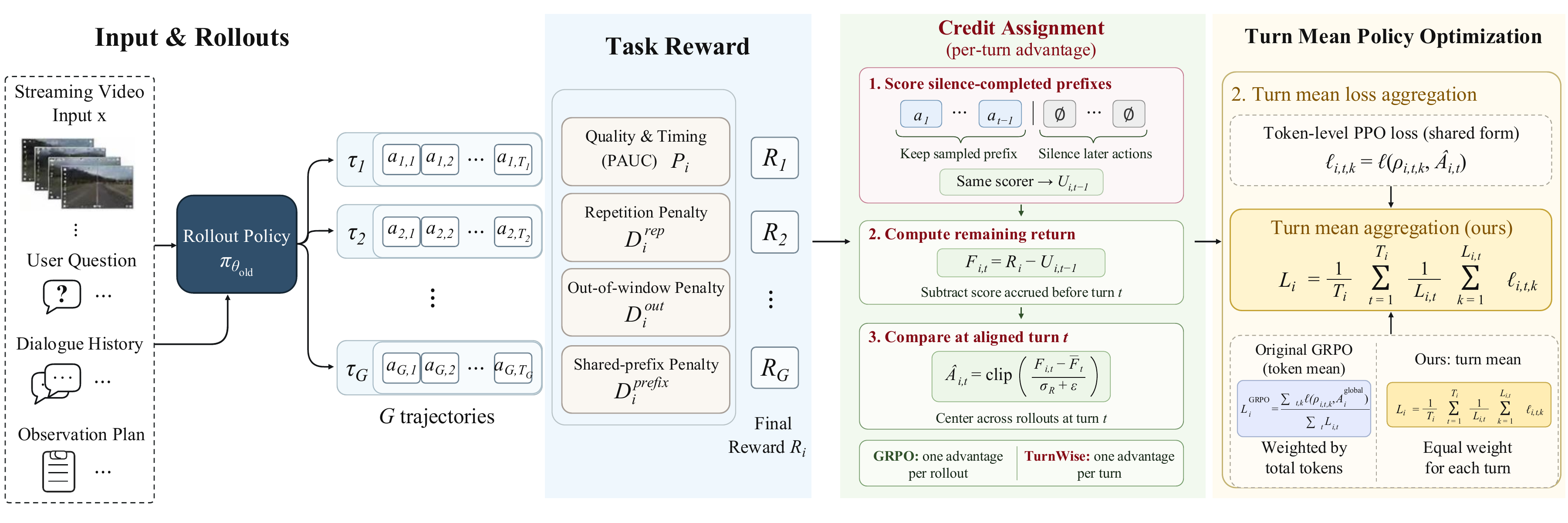}
    \caption{\textbf{Overview of TurnWise.} Given streaming video input $x$, user queries, dialogue history, and an observation plan, the policy samples a group of trajectories $\{\tau_i\}_{i=1}^{G}$. TurnWise converts trajectory rewards $R_i$ into group-relative remaining-return advantages $\hat A_{i,t}$ through silence-completed prefix scoring. Turn-mean aggregation then gives every response-or-wait decision equal weight in the trajectory loss $L_i$.}
    \label{fig:overview}
\end{figure}

\Cref{fig:overview} presents an overview of TurnWise.
TurnWise aligns policy optimization with streaming decisions through two coupled mechanisms: group-relative remaining-return advantages and equal weighting of decision turns.

\subsection{Streaming Decisions and Trajectory Rewards}
\label{sec:method:setting}

A streaming instance comprises video input $x$, user queries, initial dialogue history $h_0$, and an observation plan specifying the decision-time grid.
At decision turn $t$, the policy's context $h_{i,t}$ contains only the available video-text prefix and its previously generated actions in sampled trajectory $i$.
The policy $\pi_\theta$ generates either a textual response or the explicit waiting output \texttt{NO REPLY}, denoted by $\varnothing$.
Waiting produces no user-visible answer, but remains a generated action optimized alongside responses.
To learn these decisions, we sample $G$ trajectories $\tau_i=(a_{i,1},\ldots,a_{i,T_i})$ from the rollout policy, with $a_{i,t}\sim\pi_{\theta_{\mathrm{old}}}(\cdot\mid h_{i,t})$.
Here, $T_i$ denotes the number of decision turns in trajectory $i$; aligned trajectories within a group follow the same observation plan and have a common horizon, while different instances may have different horizons.
Their observations are shared, but their dialogue histories can differ through earlier sampled responses.
Consequently, the value of a new response depends on both the evidence available so far and the information already communicated.

Following MMDuet2~\cite{wang2025mmduet2}, we evaluate each complete trajectory with a PAUC-based reward:
\begin{equation}
    R_i=\mathcal{R}(\tau_i)
    =3P_i-2D_i^{\mathrm{rep}}-0.5D_i^{\mathrm{out}}-2D_i^{\mathrm{prefix}}.
    \label{eq:trajectory-reward}
\end{equation}
Here, $P_i$ measures response quality and timeliness within annotated reply windows~\cite{wang2025proactivevideoqa}.
For a given quality improvement, responding earlier increases its contribution over the window, linking answer content to response timing.
The remaining terms penalize unwanted response behaviors: $D_i^{\mathrm{rep}}$ is the fraction of non-silent replies whose information is already covered by previous replies; $D_i^{\mathrm{out}}$ is the fraction of out-of-window decision turns at which the model responds; and $D_i^{\mathrm{prefix}}$ counts excessively long shared-prefix violations normalized by $T_i$.
Empty denominators contribute zero, and repetition checks include non-silent assistant replies in the actual initial dialogue.
These response-level assessments ultimately produce a single interaction score.
A fixed trajectory scorer makes credit assignment and loss aggregation the controlled optimization variables.

GRPO~\cite{shao2024deepseekmath} converts this score into one group-relative advantage per trajectory:
\begin{equation}
    A_i^{\mathrm{global}}
    =\frac{R_i-\bar R}{\sigma_R+\epsilon},
    \qquad \bar R=\frac{1}{G}\sum_{j=1}^{G}R_j,
    \label{eq:global-advantage}
\end{equation}
where $\sigma_R$ is the sample standard deviation of terminal rewards within the group.
Broadcasting $A_i^{\mathrm{global}}$ to every action token expresses a preference over whole trajectories, but does not distinguish rewards earned before a turn from gains obtained from that turn onward.
Later decisions can therefore inherit credit from earlier successes or failures.
TurnWise resolves this temporal mismatch through group-relative remaining returns.

\subsection{Turn-Level Credit Assignment}
\label{sec:method:credit}

We exploit the sequential observation stream and the explicit waiting action to refine GRPO's shared trajectory advantage into turn-level credit without a critic model.
Streaming observations advance according to the shared observation plan, while the assistant may respond or remain silent at each decision point.
This structure lets us score what a trajectory prefix achieves when the stream continues but no further replies are made, providing an accrued-reward reference directly from the trajectory scorer rather than a learned state-value function.
Specifically, we retain the first $t$ sampled actions and set all later actions to silence:
\begin{equation}
    U_{i,t}=\mathcal{R}\bigl(
    a_{i,1},\ldots,a_{i,t},
    \underbrace{\varnothing,\ldots,\varnothing}_{T_i-t}
    \bigr),\qquad 0\leq t\leq T_i.
    \label{eq:prefix-score}
\end{equation}
Every prefix is scored with the same initial history, decision timestamps, reply windows, and evaluation horizon as the full trajectory.
Thus, $U_{i,t}$ measures what the retained decisions achieve if no further replies are made.
At the endpoints, $U_{i,0}$ is the shared, not necessarily zero, all-silent reference and $U_{i,T_i}=R_i$.

The score differences $r_{i,t}=U_{i,t}-U_{i,t-1}$ give the turn-level reward increments shown in \cref{fig:overview}. These increments exactly decompose the trajectory score relative to the all-silent reference.
For decision $t$, $C_{i,t-1}=U_{i,t-1}-U_{i,0}$ measures reward already accrued, while $F_{i,t}=R_i-U_{i,t-1}$ measures the additional return from the sampled continuation starting at that turn.
Using $F_{i,t}$ instead of $R_i$ excludes earlier rewards from the current decision's credit, following the return-to-go principle~\cite{sutton2018reinforcement}.
The telescoping derivation and its applicability to non-additive reward terms are given in \cref{sec:suppl:credit}.
We compare these remaining returns within each rollout group:
\begin{equation}
    \begin{aligned}
        \hat A_{i,t}
        &=\left[\frac{F_{i,t}-\bar F_t}{\sigma_R+\epsilon}
            \right]_{-a_{\max}}^{a_{\max}}\\
        &=\left[A_i^{\mathrm{global}}
        -\frac{C_{i,t-1}-\bar C_{t-1}}{\sigma_R+\epsilon}
            \right]_{-a_{\max}}^{a_{\max}}.
    \end{aligned}
    \label{eq:turnwise-advantage}
\end{equation}
Here, $\bar F_t=G^{-1}\sum_j F_{j,t}$ and $\bar C_{t-1}=G^{-1}\sum_j C_{j,t-1}$.
The operator $[z]_l^u=\min\{u,\max\{l,z\}\}$ clips the normalized advantage to $[-a_{\max},a_{\max}]$.
Here, $a_{\max}>0$ is a fixed clipping threshold ($a_{\max}=5$ in our experiments), limiting the influence of unusually large advantage magnitudes on the policy update.
The second equality makes the credit correction explicit: we subtract the group-relative reward already accrued before the current decision (see \cref{sec:suppl:credit:advantage} for the derivation).
Above-average early success no longer directly boosts later redundant replies, while below-average early performance no longer directly depresses useful follow-ups.
In particular, a trajectory with $R_i>\bar R$ can receive a negative later-turn advantage $\hat A_{i,t}$ if $F_{i,t}<\bar F_t$.
Conversely, a trajectory below the terminal-reward average can receive positive credit at a later turn when its remaining return exceeds the group average.
The first-turn credit correction is zero since $C_{i,0}=0$, and retaining $\sigma_R$ preserves a common GRPO normalization scale across turns.

To control how strongly earlier rewards are removed, we introduce $\alpha\in[0,1]$ to scale the accrued-reward correction before clipping:
\begin{equation}
    \hat A_{i,t}^{(\alpha)}
    =\left[A_i^{\mathrm{global}}
        -\alpha\frac{C_{i,t-1}-\bar C_{t-1}}{\sigma_R+\epsilon}
        \right]_{-a_{\max}}^{a_{\max}}.
    \label{eq:alpha-advantage}
\end{equation}
This is equivalent to interpolating between the global and remaining-return advantages before clipping.
At $\alpha=0$, all turns share the clipped GRPO advantage; at $\alpha=1$, $\hat A_{i,t}^{(1)}=\hat A_{i,t}$ is the fully corrected turn-specific advantage in \cref{eq:turnwise-advantage}.
As detailed in \cref{sec:experiments:training}, we linearly increase $\alpha$ from 0 to 1 over the first $10\%$ of scheduled updates and then keep it at 1, gradually transitioning from shared to turn-specific credit.

These advantages reuse the sampled trajectory's quality and repetition evaluations: a silent suffix changes neither a retained response nor its preceding replies.
A deterministic scan updates prefix scores and their denominators, yielding critic-free turn-level advantages without additional rollouts or judge calls.
Each resulting advantage is assigned to the generated action tokens within its turn.

\subsection{Turn-Mean Policy Optimization}
\label{sec:method:loss}

Within streaming GRPO, turn-level advantages assign credit to individual decisions, but token-mean loss aggregation still weights these decisions by response length.
Let $L_{i,t}$ denote the number of generated action tokens at turn $t$.
Under token-mean aggregation, this turn receives weight $L_{i,t}/\sum_s L_{i,s}$, so a long response outweighs a short waiting action.
We introduce turn-mean policy optimization to give each streaming decision equal aggregation weight regardless of its output length.

Let $\rho_{i,t,k}(\theta)$ be the current-to-rollout policy probability ratio for token $a_{i,t,k}$, conditioned on the turn history and preceding action tokens.
GRPO optimizes generated tokens with a clipped probability-ratio surrogate~\cite{shao2024deepseekmath,schulman2017ppo}:
\begin{equation}
    \ell_{\mathrm{clip}}(\rho,A)
    =-\min\bigl\{\rho A,\,[\rho]_{1-\delta}^{1+\delta}A\bigr\},
    \label{eq:token-surrogate}
\end{equation}
where $\delta$ controls probability-ratio clipping.
Let $\ell(\rho,A)$ denote this surrogate with the dual-clipping safeguard detailed in \cref{sec:suppl:policy}.
TurnWise applies the corrected group-relative advantage to each token in its turn, giving $\ell_{i,t,k}(\theta)=\ell(\rho_{i,t,k}(\theta),\hat A_{i,t}^{(\alpha)})$.

As shown in \cref{fig:overview}, we first average token losses within each turn and then average across turns:
\begin{equation}
    L_i(\theta)
    =\frac{1}{T_i}
    \sum_{t=1}^{T_i}\frac{1}{L_{i,t}}
    \sum_{k=1}^{L_{i,t}}\ell_{i,t,k}(\theta),
    \label{eq:turn-mean-loss}
\end{equation}
Each response-or-wait decision thus receives weight $1/T_i$, preventing response length from determining its share of the policy loss.
\Cref{alg:turnwise} applies this turn-mean aggregation during each GRPO update.

\subsection{Unified TurnWise Objective}
\label{sec:method:summary}

The TurnWise objective couples turn-specific remaining-return advantages with length-independent decision weights.
GRPO first normalizes $R_i$ within each rollout group to obtain the shared advantage $A_i^{\mathrm{global}}$ in \cref{eq:global-advantage}.
Without KL regularization, the trajectory-averaged GRPO loss over $B$ sampled trajectories is
\begin{equation}
    \mathcal L_{\mathrm{GRPO}}(\theta)
    =\sum_{i,t,k}
    \frac{\ell\bigl(\rho_{i,t,k}(\theta),A_i^{\mathrm{global}}\bigr)}
         {B\sum_{s=1}^{T_i}L_{i,s}}.
    \label{eq:grpo-loss}
\end{equation}
Each token therefore uses the same trajectory-level credit, and each turn's contribution scales with its output length.

We correct the shared advantage using \cref{eq:alpha-advantage} and adopt turn-mean aggregation from \cref{eq:turn-mean-loss}, yielding the unified objective:
\begin{equation}
    \begin{aligned}
        \hat A_{i,t}^{(\alpha)}
        &=\left[A_i^{\mathrm{global}}
        -\alpha\frac{C_{i,t-1}-\bar C_{t-1}}{\sigma_R+\epsilon}
        \right]_{-a_{\max}}^{a_{\max}},\\
        \mathcal L_{\mathrm{TW}}(\theta)
        &=\sum_{i,t,k}
        \frac{\ell\bigl(\rho_{i,t,k}(\theta),\hat A_{i,t}^{(\alpha)}\bigr)}
             {BT_iL_{i,t}}.
    \end{aligned}
    \label{eq:turnwise-final}
\end{equation}
The advantage correction removes reward accrued before the current decision, while the denominator $BT_iL_{i,t}$ makes decision weights independent of response length.
With full correction $\alpha=1$, the policy learns from group-relative remaining returns at the response-or-wait granularity.
\Cref{alg:turnwise} combines group sampling, trajectory scoring, turn-level advantage correction, and turn-mean policy updates in this GRPO training loop.
\begin{algorithm}[t]
    \caption{GRPO-Style TurnWise Training}
    \label{alg:turnwise}
    \small
    \begin{algorithmic}[1]
        \Require Streaming instances $\mathcal D$, policy $\pi_\theta$, trajectory scorer $\mathcal R$, rollout group size $G$, GRPO updates $S$, learning rate $\eta$.
        \Ensure Trained streaming policy $\pi_\theta$.
        \State Initialize AdamW with learning rate $\eta$
        \For{GRPO iteration $s=0,\ldots,S-1$}
            \State $\theta_{\mathrm{old}}\gets\theta$
            \State $\alpha\gets\min\{1,s/\lceil0.1S\rceil\}$ \Comment{Credit warm-up}
            \State Sample a batch of streaming instances from $\mathcal D$
            \ForAll{streaming instances in the batch}
                \State Sample aligned $\{\tau_i\}_{i=1}^{G}$ with $\pi_{\theta_{\mathrm{old}}}$ \Comment{Respond or wait}
                \State Evaluate $R_i=\mathcal R(\tau_i)$ and cache reply scores
                \State Compute $\{U_{i,t}\}$ via \cref{eq:prefix-score} \Comment{Cached scan}
                \State Compute $\bar R$ and $\sigma_R$ from $\{R_i\}_{i=1}^{G}$
                \If{$\sigma_R=0$}
                    \State $\hat A_{i,t}^{(\alpha)}\gets0$ for all $i,t$ in the group
                \Else
                    \State Compute $A_i^{\mathrm{global}}$ via \cref{eq:global-advantage}
                    \For{each aligned decision turn $t$}
                        \State $F_{i,t}\gets R_i-U_{i,t-1}$ for all $i$
                        \State $\bar F_t\gets G^{-1}\sum_{j=1}^{G}F_{j,t}$
                        \State Compute $\hat A_{i,t}^{(\alpha)}$ via \cref{eq:alpha-advantage}
                    \EndFor
                \EndIf
            \EndFor
            \State Compute token ratios $\rho_{i,t,k}(\theta)$ for the batch
            \State Compute $\ell_{i,t,k}$ with turn advantage $\hat A_{i,t}^{(\alpha)}$
            \State Compute $L_i$ via \cref{eq:turn-mean-loss} \Comment{Turn-mean loss}
            \State $B\gets$ total number of sampled trajectories
            \State Compute $\mathcal L_{\mathrm{TW}}$ via \cref{eq:turnwise-final}
            \State Update $\theta$ using AdamW and $\nabla_\theta\mathcal L_{\mathrm{TW}}$
        \EndFor
    \end{algorithmic}
\end{algorithm}

\section{Experiments}
\label{sec:experiments}
\begin{table}[tbp]
    \centering
    \caption{TurnWise achieves the best aggregate scores on ProactiveVQA, OVO-Timing, and StreamingBench with both backbones. Within each backbone, the best and second-best results are \textbf{bold} and \underline{underlined}, respectively. $\Delta$ denotes TurnWise minus the strongest non-TurnWise result in each aggregate column.}
    \label{tab:main-results}
    \begingroup
    \small
    \setlength{\tabcolsep}{5pt}
    \renewcommand{\arraystretch}{1.10}
    \definecolor{mainresultssection}{HTML}{F1F1F1}
    \definecolor{mainresultsours}{HTML}{FDEBEC}
    \definecolor{mainresultsgain}{HTML}{008000}
    \definecolor{mainresultsloss}{HTML}{C52323}
    \definecolor{mainresultsseparator}{HTML}{D8D8D8}
    \resizebox{\linewidth}{!}{%
    \begin{tabular}{l*{12}{r}}
        \toprule
        \multirow{3}{*}{\textbf{Method}} & \multicolumn{5}{c}{\textbf{ProactiveVQA}} & \multicolumn{4}{c}{\textbf{OVO-Timing}} & \multicolumn{3}{c}{\textbf{StreamingBench}} \\
        \cmidrule(lr){2-6}\cmidrule(lr){7-10}\cmidrule(lr){11-13}
         & \multicolumn{5}{c}{\textbf{PAUC $\uparrow$}} & \multicolumn{4}{c}{\textbf{Timing F1 $\uparrow$}} & \multirow{2}{*}{\textbf{\shortstack{RealTime\\Acc. $\uparrow$}}} & \multirow{2}{*}{\textbf{\shortstack{Proactive\\Acc. $\uparrow$}}} & \multirow{2}{*}{\textbf{Avg. $\uparrow$}} \\
        \cmidrule(lr){2-6}\cmidrule(lr){7-10}
         & \textbf{Ego} & \textbf{TV} & \textbf{VAD} & \textbf{Web} & \textbf{Overall} & \textbf{CRR} & \textbf{SSR} & \textbf{REC} & \textbf{Avg.} & & & \\
        \midrule
        \rowcolor{mainresultssection}
        \multicolumn{13}{l}{\textit{Qwen2.5-VL-3B}} \\
        Baseline & 9.32 & 27.85 & 8.98 & 42.45 & 22.15 & 0.00 & 0.53 & 3.46 & 1.33 & 69.42 & 22.40 & 45.91 \\
        $+$ SFT & 26.05 & 27.60 & 26.30 & 35.00 & 28.74 & 0.00 & 0.95 & 0.76 & 0.57 & 74.68 & \underline{22.80} & \underline{48.74} \\
        \arrayrulecolor{mainresultsseparator}\midrule[0.35pt]\arrayrulecolor{black}
        $+$ MMDuet2 & \underline{44.10} & \underline{39.70} & \underline{28.70} & 45.90 & \underline{39.60} & \underline{13.86} & \textbf{17.56} & 21.76 & \underline{17.72} & 74.88 & 20.40 & 47.64 \\
        $+$ ThinkStream & 32.90 & 34.48 & 15.72 & \underline{55.43} & 34.63 & 4.09 & 4.54 & 15.60 & 8.08 & 72.52 & 16.40 & 44.46 \\
        $+$ StreamPro & 39.73 & 37.96 & 22.15 & 52.29 & 38.03 & 7.98 & 6.49 & \underline{29.02} & 14.50 & \underline{75.16} & 20.40 & 47.78 \\
        \rowcolor{mainresultsours}
        $+$ \textbf{TurnWise} & \textbf{44.69} & \textbf{43.01} & \textbf{29.18} & \textbf{57.36} & \textbf{43.56} & \textbf{18.02} & \underline{14.61} & \textbf{32.15} & \textbf{21.59} & \textbf{75.85} & \textbf{23.00} & \textbf{49.43} \\
        $\Delta$ & & & & & \textcolor{mainresultsgain}{$+3.96$} & & & & \textcolor{mainresultsgain}{$+3.87$} & & & \textcolor{mainresultsgain}{$+0.69$} \\
        \midrule
        \rowcolor{mainresultssection}
        \multicolumn{13}{l}{\textit{Qwen3-VL-4B}} \\
        Baseline & 24.00 & 31.80 & 15.09 & 48.42 & 29.83 & 4.21 & 4.56 & \underline{16.66} & 8.48 & 73.57 & 32.80 & 53.19 \\
        $+$ SFT & 29.10 & 27.07 & \underline{28.46} & 36.33 & 30.24 & 2.78 & 0.95 & 1.06 & 1.60 & 77.56 & \underline{35.60} & \underline{56.58} \\
        \arrayrulecolor{mainresultsseparator}\midrule[0.35pt]\arrayrulecolor{black}
        $+$ MMDuet2 & 24.30 & 28.57 & 24.95 & 42.91 & 30.18 & \textbf{11.88} & 4.25 & 11.57 & \underline{9.23} & 77.28 & 30.40 & 53.84 \\
        $+$ ThinkStream & \underline{41.29} & \textbf{38.49} & 19.47 & 45.48 & \underline{36.18} & 5.67 & \underline{6.09} & 15.09 & 8.95 & 77.04 & 34.80 & 55.92 \\
        $+$ StreamPro & 36.39 & 36.43 & 14.82 & \underline{51.02} & 34.67 & 4.34 & 4.63 & 16.52 & 8.50 & \underline{78.24} & 32.80 & 55.52 \\
        \rowcolor{mainresultsours}
        $+$ \textbf{TurnWise} & \textbf{42.63} & \underline{37.73} & \textbf{31.95} & \textbf{57.45} & \textbf{42.44} & \underline{8.68} & \textbf{7.35} & \textbf{17.43} & \textbf{11.15} & \textbf{79.52} & \textbf{38.40} & \textbf{58.96} \\
        $\Delta$ & & & & & \textcolor{mainresultsgain}{$+6.26$} & & & & \textcolor{mainresultsgain}{$+1.92$} & & & \textcolor{mainresultsgain}{$+2.38$} \\
        \bottomrule
    \end{tabular}%
    }
    \endgroup
\end{table}

Our evaluation measures benchmark performance, the complementary effects of credit assignment and loss aggregation, and the resulting changes in response behavior.

\subsection{Training Pipeline}
\label{sec:experiments:training}

We first fine-tune Qwen2.5-VL-3B~\cite{bai2025qwen25vl} and Qwen3-VL-4B~\cite{bai2025qwen3vl} on EgoExo4D~\cite{grauman2024egoexo4d}, EgoExoLearn~\cite{huang2024egoexolearn}, Live-WhisperX~\cite{chen2025live}, Tarsier2~\cite{yuan2025tarsier2}, and LLaVA-Video~\cite{zhang2024llavavideo}.
Both models use the same SFT corpus of 100,874 valid instruction examples, with $1\%$ held out for validation.
SFT runs for one epoch at learning rate $10^{-5}$ with the vision tower frozen.
We then optimize the streaming policies with GRPO-style reinforcement learning~\cite{shao2024deepseekmath}, using groups of sampled trajectories to form relative advantage signals.
RL uses 3,983 streaming instances with 20-second target segments, initial dialogue context, and at most 90 frames sampled every two seconds, following MMDuet2~\cite{wang2025mmduet2}.
We use the PAUC-based reward in \cref{eq:trajectory-reward}, judged by \texttt{gpt-oss-20b}~\cite{openai2025gptoss}.

All RL configurations use batch size 8 and $G=4$ rollouts per instance, with temperature 1.2 and top-$k=10$, following MMDuet2~\cite{wang2025mmduet2}.
We train for one epoch with AdamW at learning rate $10^{-6}$, one policy optimization epoch per batch, and no KL or entropy regularization.
TurnWise freezes the vision tower and uses turn-mean aggregation, with $\alpha$ linearly warmed up to 1 over the first $10\%$ of scheduled updates.
We set $\epsilon=10^{-6}$, $a_{\max}=5$, $\delta=0.2$, and $c=3$.

\subsection{Benchmarks}
\label{sec:experiments:benchmarks}

We evaluate on ProactiveVQA~\cite{wang2025proactivevideoqa}, using 300 instances across Ego, TV, VAD, and Web to assess answer quality and timeliness within reply windows.
OVO-Timing~\cite{niu2025ovobench} evaluates event-triggered responses on CRR, SSR, and REC.
StreamingBench~\cite{lin2026streamingbench} tests answer correctness from observed video prefixes through Real-Time Visual Understanding, and joint answer correctness and timing through Proactive Output.
Together, these benchmarks assess complementary aspects of streaming policies: proactive answer quality, event-triggered timing, and real-time visual understanding.

\subsection{Evaluations}
\label{sec:experiments:evaluation}

We compare TurnWise with SFT models and the RL baselines MMDuet2, ThinkStream~\cite{liu2026thinkstream}, and StreamPro~\cite{li2026streampro}, implemented in the same streaming GRPO framework.
All models use greedy decoding with observed video prefixes and their own previous responses as context.

On ProactiveVQA, we report PAUC with $\omega=0.5$, judged by \texttt{GPT-5.6-luna} and averaged equally across four categories; ablations also report repetition rates among non-silent replies.
OVO-Timing uses Timing F1 with one-to-one matching within $\pm2$ seconds.
We compute F1 from mean per-instance precision and recall for each task, then average equally across CRR, SSR, and REC.
For StreamingBench, we report multiple-choice accuracy on Real-Time Visual Understanding and joint answer-and-timing accuracy within $\pm2$ seconds on Proactive Output, and their equal-weight mean.

\section{Results and Discussions}
\label{sec:results}

\subsection{Main Results}
\label{sec:results:main}

TurnWise achieves the best aggregate scores on all three benchmarks with both backbones.
Component analysis shows that combining turn-level credit with turn-mean aggregation improves task performance while reducing redundant replies.

\begin{figure}[t]
    \centering
    \includegraphics[width=\textwidth]{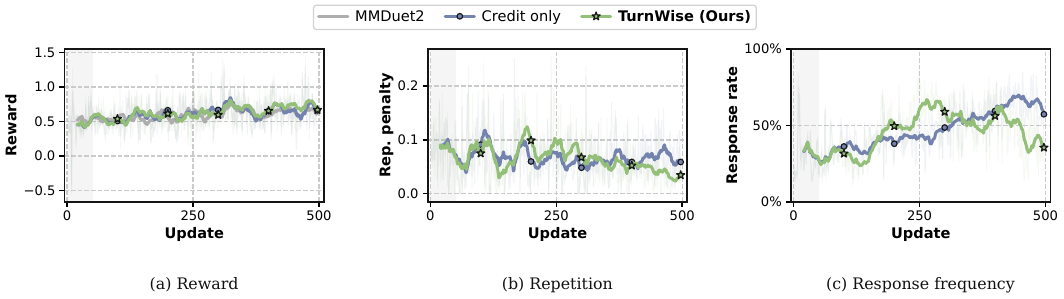}
    \caption{\textbf{TurnWise achieves higher late-stage reward with fewer redundant replies relative to credit-only optimization.} Panel a shows reward, panel b shows reply-normalized repetition penalty magnitude, and panel c shows non-silent decision fraction. Faint traces are batch means. Solid curves are trailing 20-update averages. Gray shading marks credit warmup.}
    \label{fig:training-dynamics}
\end{figure}

\begin{table}[tbp]
    \centering
    \begingroup
    \renewenvironment{table}[1][]{\captionsetup{type=table}}{}
    \begin{minipage}[t]{0.49\textwidth}
\begin{table}[t]
    \centering
    \caption{Complementary effects of turn-level credit and turn-mean aggregation on Qwen2.5-VL-3B. Their combination achieves the highest task scores and reduces repetition relative to either component alone. PAUC, F1, and Avg. are benchmark-level scores; Rep. measures repetition among non-silent replies. The best and second-best results are \textbf{bold} and \underline{underlined}.}
    \label{tab:component-ablation}
    \begingroup
    \small
    \setlength{\tabcolsep}{1.2pt}
    \renewcommand{\arraystretch}{1.12}
    \definecolor{componentours}{HTML}{FDEBEC}
    \resizebox{\linewidth}{!}{%
    \begin{tabular}{@{}lccrrrr@{}}
        \toprule
        \multirow{2}{*}{\textbf{Method}} & \multirow{2}{*}{\textbf{\shortstack{Credit\\Assign.}}} & \multirow{2}{*}{\textbf{\shortstack{Turn\\Mean}}} & \multicolumn{2}{c}{\textbf{\shortstack{Proactive\\VQA}}} & \textbf{\shortstack{OVO-\\Timing}} & \textbf{\shortstack{Streaming\\Bench}} \\
        \cmidrule(lr){4-5}\cmidrule(lr){6-6}\cmidrule(lr){7-7}
         & & & \textbf{PAUC $\uparrow$} & \textbf{Rep. $\downarrow$} & \textbf{F1 $\uparrow$} & \textbf{Avg. $\uparrow$} \\
        \midrule
        Base & --- & --- & 22.15 & 65.40 & 1.33 & 45.91 \\
        SFT & --- & --- & 28.74 & \textbf{3.79} & 0.57 & 48.74 \\
        \midrule
        MMDuet2 & --- & --- & 39.60 & 6.08 & 17.72 & 47.64 \\
        $+$ Turn Mean & --- & $\checkmark$ & 41.87 & 9.78 & 8.17 & 47.58 \\
        $+$ Credit & $\checkmark$ & --- & \underline{43.12} & 15.53 & \underline{18.63} & \underline{48.94} \\
        \rowcolor{componentours}
        \textbf{TurnWise} & $\checkmark$ & $\checkmark$ & \textbf{43.56} & \underline{5.98} & \textbf{21.59} & \textbf{49.43} \\
        \bottomrule
    \end{tabular}%
    }
    \endgroup
\end{table}

    \end{minipage}\hfill
    \begin{minipage}[t]{0.49\textwidth}
\begin{table}[t]
    \centering
    \caption{With turn-mean aggregation, $\alpha=1$ is best across all four metrics on Qwen2.5-VL-3B. Metrics follow \cref{tab:component-ablation}; \textbf{bold}/\underline{underline} mark best/second-best within each configuration.}
    \label{tab:alpha-sensitivity}
    \begingroup
    \small
    \setlength{\tabcolsep}{11pt}
    \renewcommand{\arraystretch}{1.12}
    \definecolor{alphasection}{HTML}{F1F1F1}
    \definecolor{alphaours}{HTML}{FDEBEC}
    \resizebox{\linewidth}{!}{%
    \begin{tabular}{@{}crrrr@{}}
        \toprule
        \multirow{2}{*}{\textbf{$\alpha$}} & \multicolumn{2}{c}{\textbf{\shortstack{Proactive\\VQA}}} & \textbf{\shortstack{OVO-\\Timing}} & \textbf{\shortstack{Streaming\\Bench}} \\
        \cmidrule(lr){2-3}\cmidrule(lr){4-4}\cmidrule(lr){5-5}
         & \textbf{PAUC $\uparrow$} & \textbf{Rep. $\downarrow$} & \textbf{F1 $\uparrow$} & \textbf{Avg. $\uparrow$} \\
        \midrule
        \rowcolor{alphasection}
        \multicolumn{5}{l}{\textit{Credit Assignment}} \\
        0.25 & \textbf{44.50} & \underline{16.81} & \textbf{19.51} & 47.46 \\
        0.50 & 43.34 & 18.91 & \underline{19.41} & \underline{48.30} \\
        0.75 & \underline{44.24} & 25.16 & 12.97 & 46.84 \\
        1.00 & 43.12 & \textbf{15.53} & 18.63 & \textbf{48.94} \\
        \midrule
        \rowcolor{alphasection}
        \multicolumn{5}{l}{\textit{Credit Assignment + Turn Mean}} \\
        0.25 & 41.37 & \underline{12.49} & \underline{16.68} & \underline{48.05} \\
        0.50 & 41.71 & 36.04 & 10.53 & 46.44 \\
        0.75 & \underline{42.00} & 12.95 & 9.73 & 46.95 \\
        \rowcolor{alphaours}
        \textbf{1.00} & \textbf{43.56} & \textbf{5.98} & \textbf{21.59} & \textbf{49.43} \\
        \bottomrule
    \end{tabular}%
    }
    \endgroup
\end{table}

    \end{minipage}
    \endgroup
\end{table}

\begin{figure}[t]
    \centering
    \includegraphics[width=\textwidth]{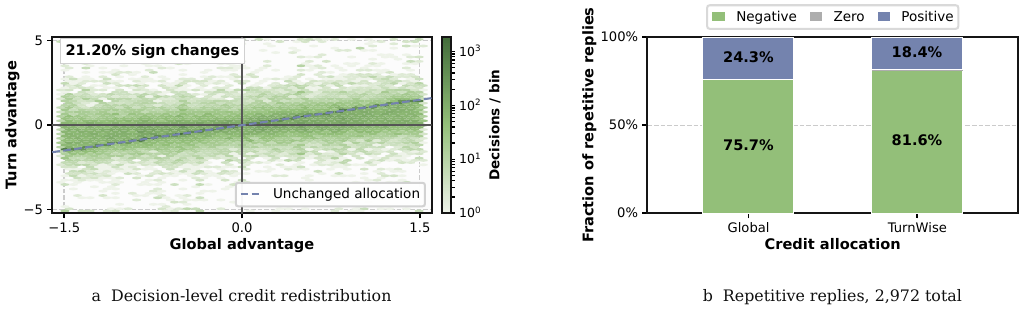}
    \caption{\textbf{Turn-level credit assigns negative advantages to $59.83\%$ of repetitive replies with positive global advantages (432 of 722).} Both allocations use the same sampled groups from Qwen2.5-VL-3B training, using complete trajectories with full credit correction at $\alpha=1$. Panel a shows advantage density over 84,184 decisions. Turn advantages are clipped to $\pm5$ as in training. The dashed diagonal marks unchanged allocation. Panel b shows advantage signs for all 2,972 Judge-labeled repetitive replies in this cohort.}
    \label{fig:credit-allocation}
\end{figure}

\paragraph{Benchmark Results.}
\Cref{tab:main-results} compares TurnWise with SFT and GRPO-based RL baselines on three streaming benchmarks.
TurnWise achieves the best aggregate scores on three benchmarks with both backbones.
On Qwen2.5-VL-3B, it reaches 43.56 PAUC on ProactiveVQA, 21.59 average Timing F1 on OVO-Timing, and 49.43 average accuracy on StreamingBench, exceeding the strongest non-TurnWise results by 3.96, 3.87, and 0.69 points, respectively.
On Qwen3-VL-4B, the scores are 42.44, 11.15, and 58.96, with gains of 6.26, 1.92, and 2.38 points.

The improvements extend across both proactive interaction and visual understanding.
On ProactiveVQA, TurnWise leads all four categories with the 3B backbone and three of four with the 4B backbone.
For the latter, VAD and Web improve by 3.49 and 6.43 points over the strongest competing results.
On StreamingBench, TurnWise obtains the highest real-time and proactive accuracies with both backbones.
In particular, the 4B model reaches 38.40 proactive accuracy, compared with 35.60 for the strongest baseline, while also improving real-time accuracy from 78.24 to 79.52.
Across both backbones, TurnWise improves proactive interaction, response timing, and real-time visual understanding.

\paragraph{Complementary Effects of Credit Assignment and Turn-Mean Aggregation.}
On Qwen2.5-VL-3B, combining turn-level credit with turn-mean aggregation reduces repetition by $61.5\%$ relative to credit-only optimization, from 15.53 to 5.98, while increasing Timing F1 by 2.96 points and StreamingBench average accuracy by 0.49 points (\cref{tab:component-ablation}).
Turn-level credit determines each decision's learning direction; turn-mean aggregation determines its weight in the policy update.
Their combination achieves the highest aggregate task scores among the evaluated component configurations.
Relative to turn-mean aggregation alone, combining both components improves Timing F1 by 13.42 points and reduces repetition by $38.9\%$.

\enlargethispage{\baselineskip}
\paragraph{Alpha Sensitivity.}
Within the complete TurnWise framework on Qwen2.5-VL-3B, full correction ($\alpha=1$) achieves the highest PAUC, Timing F1, and StreamingBench accuracy and the lowest repetition among the tested correction strengths (\cref{tab:alpha-sensitivity}).
This identifies full accrued-reward correction as the operating point for TurnWise.

\subsection{Training Dynamics and Response Behavior}
\label{sec:results:analysis}

Turn-mean aggregation raises mean reward from 0.657 to 0.711 over the final 50 updates while reducing repetition-penalty magnitude from 0.069 to 0.036 (\cref{fig:training-dynamics}, panels a and b).
Both configurations use the same turn-level credit rule and Qwen2.5-VL-3B SFT initialization across 497 updates.
The outside-window reply fraction decreases from 0.537 to 0.289, while the non-silent decision fraction falls from $62.2\%$ to $37.0\%$ (panel c).
Higher reward accompanies more selective response behavior under turn-mean aggregation.

\subsection{Credit Redistribution on Real Training Trajectories}
\label{sec:results:credit-allocation}

Turn-level credit assigns negative advantages to $59.83\%$ of the repetitive replies that receive positive global advantages---432 of 722 replies in the analyzed cohort.
\Cref{fig:credit-allocation} compares the logged turn advantages with the global advantages computed from the same sampled groups and terminal rewards.
The analysis covers 11,116 complete Qwen2.5-VL-3B training trajectories with full credit correction at $\alpha=1$, comprising 84,184 decision turns.
As shown in panel a of \cref{fig:credit-allocation}, $21.20\%$ of decisions change advantage sign relative to global allocation, and $53.15\%$ of trajectories receive both positive and negative turn advantages.
Thus, a single trajectory provides distinct learning directions for its constituent decisions.

This differentiation also targets redundant replies within positively ranked trajectories.
Panel b of \cref{fig:credit-allocation} shows that turn-level credit increases the share receiving negative advantages from $75.7\%$ to $81.6\%$ across all 2,972 Judge-labeled repetitive replies in this cohort.
These recorded advantages show that turn-level credit assigns distinct learning directions within a trajectory and provides suppressive signals for redundant replies in positively ranked trajectories.

\subsection{Discussion}
\label{sec:results:discussion}

\Cref{tab:component-ablation,tab:alpha-sensitivity,fig:training-dynamics,fig:credit-allocation} highlight complementary roles for credit assignment and loss aggregation: the former determines each turn's learning signal, while the latter controls its optimization weight.
Remaining-return advantages improve credit assignment, while turn-mean aggregation balances response and waiting decisions.
Their combination improves task scores and Timing F1 while reducing redundant and outside-window replies.
TurnWise unifies credit assignment and policy optimization at the decision-turn level.
A deterministic scan of cached PAUC assessments computes the prefix references without additional rollouts or judge calls.
Across the evaluated settings, this alignment improves task performance and response timing while reducing redundant replies.

\section{Conclusion}
\label{sec:conclusion}

\begingroup
\setlength{\emergencystretch}{2em}
TurnWise makes the decision turn the common unit of credit assignment and policy optimization in streaming video RL.
Silence-completed prefix references remove previously accrued rewards from later-turn credit, while turn-mean aggregation gives response and waiting decisions equal weight, without a learned critic or additional rollouts.
Across two Qwen backbones, TurnWise achieves the best aggregate results on all three evaluated benchmarks, with component analyses supporting higher-quality, better-timed, and \mbox{more selective responses}.
\par
\endgroup

\FloatBarrier
\bibliographystyle{ieeenat_fullname}
\bibliography{main}

\begin{thebibliography}{34}
\providecommand{\natexlab}[1]{#1}
\providecommand{\url}[1]{\texttt{#1}}
\expandafter\ifx\csname urlstyle\endcsname\relax
  \providecommand{\doi}[1]{doi: #1}\else
  \providecommand{\doi}{doi: \begingroup \urlstyle{rm}\Url}\fi

\bibitem[Ahn et~al.(2024)Ahn, Choi, Yu, Kang, and Choi]{ahn2024vlmrlaif}
Daechul Ahn, Yura Choi, Youngjae Yu, Dongyeop Kang, and Jonghyun Choi.
\newblock Tuning large multimodal models for videos using reinforcement learning from {AI} feedback.
\newblock In \emph{Proceedings of the 62nd Annual Meeting of the Association for Computational Linguistics (Volume 1: Long Papers)}, pages 923--940, Bangkok, Thailand, 2024. Association for Computational Linguistics.

\bibitem[Bai et~al.(2025{\natexlab{a}})Bai, Cai, Chen, Chen, Chen, Cheng, Deng, Ding, Gao, Ge, Ge, Guo, Huang, Huang, Huang, Hui, Jiang, Li, Li, Li, Li, Lin, Lin, Liu, Liu, Liu, Liu, Liu, Liu, Lu, Luo, Lv, Men, Meng, Ren, Ren, Song, Sun, Tang, Tu, Wan, Wang, Wang, Wang, Wang, Xie, Xu, Xu, Xu, Yang, Yang, Yang, Yang, Yu, Zhang, Zhang, Zhang, Zheng, Zhong, Zhou, Zhou, Zhou, Zhu, and Zhu]{bai2025qwen3vl}
Shuai Bai, Yuxuan Cai, Ruizhe Chen, Keqin Chen, Xionghui Chen, Zesen Cheng, Lianghao Deng, Wei Ding, Chang Gao, Chunjiang Ge, Wenbin Ge, Zhifang Guo, Qidong Huang, Jie Huang, Fei Huang, Binyuan Hui, Shutong Jiang, Zhaohai Li, Mingsheng Li, Mei Li, Kaixin Li, Zicheng Lin, Junyang Lin, Xuejing Liu, Jiawei Liu, Chenglong Liu, Yang Liu, Dayiheng Liu, Shixuan Liu, Dunjie Lu, Ruilin Luo, Chenxu Lv, Rui Men, Lingchen Meng, Xuancheng Ren, Xingzhang Ren, Sibo Song, Yuchong Sun, Jun Tang, Jianhong Tu, Jianqiang Wan, Peng Wang, Pengfei Wang, Qiuyue Wang, Yuxuan Wang, Tianbao Xie, Yiheng Xu, Haiyang Xu, Jin Xu, Zhibo Yang, Mingkun Yang, Jianxin Yang, An Yang, Bowen Yu, Fei Zhang, Hang Zhang, Xi Zhang, Bo Zheng, Humen Zhong, Jingren Zhou, Fan Zhou, Jing Zhou, Yuanzhi Zhu, and Ke Zhu.
\newblock {Qwen3-VL} technical report.
\newblock \emph{arXiv preprint arXiv:2511.21631}, 2025{\natexlab{a}}.

\bibitem[Bai et~al.(2025{\natexlab{b}})Bai, Chen, Liu, Wang, Ge, Song, Dang, Wang, Wang, Tang, Zhong, Zhu, Yang, Li, Wan, Wang, Ding, Fu, Xu, Ye, Zhang, Xie, Cheng, Zhang, Yang, Xu, and Lin]{bai2025qwen25vl}
Shuai Bai, Keqin Chen, Xuejing Liu, Jialin Wang, Wenbin Ge, Sibo Song, Kai Dang, Peng Wang, Shijie Wang, Jun Tang, Humen Zhong, Yuanzhi Zhu, Mingkun Yang, Zhaohai Li, Jianqiang Wan, Pengfei Wang, Wei Ding, Zheren Fu, Yiheng Xu, Jiabo Ye, Xi Zhang, Tianbao Xie, Zesen Cheng, Hang Zhang, Zhibo Yang, Haiyang Xu, and Junyang Lin.
\newblock {Qwen2.5-VL} technical report.
\newblock \emph{arXiv preprint arXiv:2502.13923}, 2025{\natexlab{b}}.

\bibitem[Chen et~al.(2024)Chen, Lv, Wu, Lin, Song, Gao, Liu, Gao, Mao, and Shou]{chen2024videollmonline}
Joya Chen, Zhaoyang Lv, Shiwei Wu, Kevin~Qinghong Lin, Chenan Song, Difei Gao, Jia-Wei Liu, Ziteng Gao, Dongxing Mao, and Mike~Zheng Shou.
\newblock {VideoLLM-online}: Online video large language model for streaming video.
\newblock In \emph{Proceedings of the IEEE/CVF Conference on Computer Vision and Pattern Recognition (CVPR)}, pages 18407--18418, 2024.

\bibitem[Chen et~al.(2025{\natexlab{a}})Chen, Zeng, Lin, Li, Ma, and Shou]{chen2025live}
Joya Chen, Ziyun Zeng, Yiqi Lin, Wei Li, Zejun Ma, and Mike~Zheng Shou.
\newblock {LIVE}: Learning video {LLM} with streaming speech transcription at scale.
\newblock In \emph{Proceedings of the IEEE/CVF Conference on Computer Vision and Pattern Recognition (CVPR)}, pages 29083--29095, 2025{\natexlab{a}}.

\bibitem[Chen et~al.(2025{\natexlab{b}})Chen, Huang, Shi, Hu, Ye, Zhu, Liu, Molchanov, Kautz, Qi, Liu, Yin, Lu, and Han]{chen2025scalingrl}
Yukang Chen, Wei Huang, Baifeng Shi, Qinghao Hu, Hanrong Ye, Ligeng Zhu, Zhijian Liu, Pavlo Molchanov, Jan Kautz, Xiaojuan Qi, Sifei Liu, Hongxu Yin, Yao Lu, and Song Han.
\newblock Scaling {RL} to long videos.
\newblock In \emph{Advances in Neural Information Processing Systems}, 2025{\natexlab{b}}.

\bibitem[Dong et~al.(2026{\natexlab{a}})Dong, Wang, Liu, Li, Hou, and Wei]{dong2026interleaved}
Shuai Dong, Siyuan Wang, Xingyu Liu, Chenglin Li, Haowen Hou, and Zhongyu Wei.
\newblock Interleaved latent visual reasoning with selective perceptual modeling.
\newblock In \emph{Proceedings of the 64th Annual Meeting of the Association for Computational Linguistics (Volume 1: Long Papers)}, pages 29316--29335, San Diego, California, United States, 2026{\natexlab{a}}. Association for Computational Linguistics.

\bibitem[Dong et~al.(2026{\natexlab{b}})Dong, Zhu, Xu, Xie, {Liuwenpu}, Wang, Tuo, Wang, Wang, Shao, Yang, Zhao, Ma, Chang, Wu, Yu, Wu, He, Chen, Wang, Tian, Chen, Liu, Wu, Guo, Hou, Chen, Wei, and Wang]{dong2026rl}
Shuai Dong, Yongfu Zhu, Yuqi Xu, Weichu Xie, {Liuwenpu}, Ziyue Wang, Kaiwen Tuo, Congcong Wang, Siyuan Wang, Wenqi Shao, Shuai Yang, Ji Zhao, Caoyuan Ma, Wenzheng Chang, Taiqiang Wu, Xinlei Yu, Hongrui Wu, Xiaoxuan He, Fangke Chen, Dianyi Wang, Kanghui Tian, Sirry Chen, Xingyu Liu, Xiangnan Wu, Jiawei Guo, Haowen Hou, LingHan Chen, Zhongyu Wei, and Jiaqi Wang.
\newblock {RL} starts before {RL}: On policy distillation for better reinforcement learning.
\newblock \emph{arXiv preprint arXiv:2609.28145}, 2026{\natexlab{b}}.

\bibitem[Feng et~al.(2025)Feng, Gong, Li, Guo, Wang, Peng, Wu, Zhang, Wang, and Yue]{feng2025videor1}
Kaituo Feng, Kaixiong Gong, Bohao Li, Zonghao Guo, Yibing Wang, Tianshuo Peng, Junfei Wu, Xiaoying Zhang, Benyou Wang, and Xiangyu Yue.
\newblock {Video-R1}: Reinforcing video reasoning in {MLLMs}.
\newblock In \emph{Advances in Neural Information Processing Systems}, 2025.

\bibitem[Grauman et~al.(2024)Grauman, Westbury, Torresani, Kitani, Malik, Afouras, Ashutosh, Baiyya, Bansal, Boote, Byrne, Chavis, Chen, Cheng, Chu, Crane, Dasgupta, Dong, Escobar, Forigua, Gebreselasie, Haresh, Huang, Islam, Jain, Khirodkar, Kukreja, Liang, Liu, Majumder, Mao, Martin, Mavroudi, Nagarajan, Ragusa, Ramakrishnan, Seminara, Somayazulu, Song, Su, Xue, Zhang, Zhang, Castillo, Chen, Fu, Furuta, Gonzalez, Gupta, Hu, Huang, Huang, Khoo, Kumar, Kuo, Lakhavani, Liu, Luo, Luo, Meredith, Miller, Oguntola, Pan, Peng, Pramanick, Ramazanova, Ryan, Shan, Somasundaram, Song, Southerland, Tateno, Wang, Wang, Yagi, Yan, Yang, Yu, Zha, Zhao, Zhao, Zhu, Zhuo, Arbelaez, Bertasius, Damen, Engel, Farinella, Furnari, Ghanem, Hoffman, Jawahar, Newcombe, Park, Rehg, Sato, Savva, Shi, Shou, and Wray]{grauman2024egoexo4d}
Kristen Grauman, Andrew Westbury, Lorenzo Torresani, Kris Kitani, Jitendra Malik, Triantafyllos Afouras, Kumar Ashutosh, Vijay Baiyya, Siddhant Bansal, Bikram Boote, Eugene Byrne, Zach Chavis, Joya Chen, Feng Cheng, Fu-Jen Chu, Sean Crane, Avijit Dasgupta, Jing Dong, Maria Escobar, Cristhian Forigua, Abrham Gebreselasie, Sanjay Haresh, Jing Huang, Md~Mohaiminul Islam, Suyog Jain, Rawal Khirodkar, Devansh Kukreja, Kevin~J Liang, Jia-Wei Liu, Sagnik Majumder, Yongsen Mao, Miguel Martin, Effrosyni Mavroudi, Tushar Nagarajan, Francesco Ragusa, Santhosh~Kumar Ramakrishnan, Luigi Seminara, Arjun Somayazulu, Yale Song, Shan Su, Zihui Xue, Edward Zhang, Jinxu Zhang, Angela Castillo, Changan Chen, Xinzhu Fu, Ryosuke Furuta, Cristina Gonzalez, Prince Gupta, Jiabo Hu, Yifei Huang, Yiming Huang, Weslie Khoo, Anush Kumar, Robert Kuo, Sach Lakhavani, Miao Liu, Mi Luo, Zhengyi Luo, Brighid Meredith, Austin Miller, Oluwatumininu Oguntola, Xiaqing Pan, Penny Peng, Shraman Pramanick, Merey Ramazanova, Fiona Ryan, Wei Shan,
  Kiran Somasundaram, Chenan Song, Audrey Southerland, Masatoshi Tateno, Huiyu Wang, Yuchen Wang, Takuma Yagi, Mingfei Yan, Xitong Yang, Zecheng Yu, Shengxin~Cindy Zha, Chen Zhao, Ziwei Zhao, Zhifan Zhu, Jeff Zhuo, Pablo Arbelaez, Gedas Bertasius, Dima Damen, Jakob Engel, Giovanni~Maria Farinella, Antonino Furnari, Bernard Ghanem, Judy Hoffman, C.V. Jawahar, Richard Newcombe, Hyun~Soo Park, James~M. Rehg, Yoichi Sato, Manolis Savva, Jianbo Shi, Mike~Zheng Shou, and Michael Wray.
\newblock {Ego-Exo4D}: Understanding skilled human activity from first- and third-person perspectives.
\newblock In \emph{Proceedings of the IEEE/CVF Conference on Computer Vision and Pattern Recognition (CVPR)}, pages 19383--19400, 2024.

\bibitem[Guo et~al.(2026)Guo, Chen, Wang, Li, Zhao, Liu, Dong, Zhai, and Zhou]{guo2026beyond}
Jiawei Guo, Yu Chen, Xiang Wang, Shuai Li, Xinpei Zhao, Huaxing Liu, Shuai Dong, Feifei Zhai, and Yu Zhou.
\newblock Beyond visual memory: Mechanistic diagnostics of latent visual reasoning.
\newblock \emph{arXiv preprint arXiv:2606.01287}, 2026.

\bibitem[Huang et~al.(2024)Huang, Chen, Xu, Zhang, Yang, Pei, Zhang, Dong, Wang, Wang, and Qiao]{huang2024egoexolearn}
Yifei Huang, Guo Chen, Jilan Xu, Mingfang Zhang, Lijin Yang, Baoqi Pei, Hongjie Zhang, Lu Dong, Yali Wang, Limin Wang, and Yu Qiao.
\newblock {EgoExoLearn}: A dataset for bridging asynchronous ego- and exo-centric view of procedural activities in real world.
\newblock In \emph{Proceedings of the IEEE/CVF Conference on Computer Vision and Pattern Recognition (CVPR)}, pages 22072--22086, 2024.

\bibitem[Li et~al.(2026)Li, Xiao, Yue, Xu, Yao, Li, Fu, Ju, Luan, and Jin]{li2026streampro}
Ao Li, Zihan Xiao, Zihao Yue, Boshen Xu, Linli Yao, Jiaze Li, Pei Fu, Jianzhong Ju, Jian Luan, and Qin Jin.
\newblock {StreamPro}: From reactive perception to proactive decision-making in streaming video.
\newblock \emph{arXiv preprint arXiv:2605.16381}, 2026.

\bibitem[Li et~al.(2025)Li, Yan, Meng, Dong, Zeng, He, Wang, Qiao, Wang, and Wang]{li2025videochatr1}
Xinhao Li, Ziang Yan, Desen Meng, Lu Dong, Xiangyu Zeng, Yinan He, Yali Wang, Yu Qiao, Yi Wang, and Limin Wang.
\newblock {VideoChat-R1}: Enhancing spatio-temporal perception via reinforcement fine-tuning.
\newblock \emph{arXiv preprint arXiv:2504.06958}, 2025.

\bibitem[Lin et~al.(2026)Lin, Fang, Chen, Cheng, Wan, Luo, Wang, Li, Liu, and Sun]{lin2026streamingbench}
Junming Lin, Zheng Fang, Chi Chen, Haoxuan Cheng, Zihao Wan, Fuwen Luo, Ziyue Wang, Peng Li, Yang Liu, and Maosong Sun.
\newblock {StreamingBench}: Assessing the gap for {MLLMs} to achieve streaming video understanding.
\newblock In \emph{ICASSP 2026 -- 2026 IEEE International Conference on Acoustics, Speech and Signal Processing (ICASSP)}, pages 12147--12151. IEEE, 2026.

\bibitem[Liu et~al.(2026)Liu, Guo, Li, Zhen, He, Ji, Ren, Zhang, Lu, and Liu]{liu2026thinkstream}
Zikang Liu, Longteng Guo, Handong Li, Ru Zhen, Xingjian He, Ruyi Ji, Xiaoming Ren, Yanhao Zhang, Haonan Lu, and Jing Liu.
\newblock Thinking in streaming video.
\newblock \emph{arXiv preprint arXiv:2603.12938}, 2026.

\bibitem[Niu et~al.(2025)Niu, Li, Miao, Ge, Zhou, He, Dong, Duan, Ding, Qian, Zhang, Zang, Cao, He, and Wang]{niu2025ovobench}
Junbo Niu, Yifei Li, Ziyang Miao, Chunjiang Ge, Yuanhang Zhou, Qihao He, Xiaoyi Dong, Haodong Duan, Shuangrui Ding, Rui Qian, Pan Zhang, Yuhang Zang, Yuhang Cao, Conghui He, and Jiaqi Wang.
\newblock {OVO-Bench}: How far is your {Video-LLMs} from real-world online video understanding?
\newblock In \emph{Proceedings of the IEEE/CVF Conference on Computer Vision and Pattern Recognition (CVPR)}, pages 18902--18913, 2025.

\bibitem[{OpenAI}(2025)]{openai2025gptoss}
{OpenAI}.
\newblock {gpt-oss-120b} \& {gpt-oss-20b} model card.
\newblock \emph{arXiv preprint arXiv:2508.10925}, 2025.

\bibitem[Qian et~al.(2025)Qian, Ding, Dong, Zhang, Zang, Cao, Lin, and Wang]{qian2025dispider}
Rui Qian, Shuangrui Ding, Xiaoyi Dong, Pan Zhang, Yuhang Zang, Yuhang Cao, Dahua Lin, and Jiaqi Wang.
\newblock {Dispider}: Enabling video {LLMs} with active real-time interaction via disentangled perception, decision, and reaction.
\newblock In \emph{Proceedings of the IEEE/CVF Conference on Computer Vision and Pattern Recognition (CVPR)}, pages 24045--24055, 2025.

\bibitem[Schulman et~al.(2017)Schulman, Wolski, Dhariwal, Radford, and Klimov]{schulman2017ppo}
John Schulman, Filip Wolski, Prafulla Dhariwal, Alec Radford, and Oleg Klimov.
\newblock Proximal policy optimization algorithms.
\newblock \emph{arXiv preprint arXiv:1707.06347}, 2017.

\bibitem[Shao et~al.(2024)Shao, Wang, Zhu, Xu, Song, Bi, Zhang, Zhang, Li, Wu, and Guo]{shao2024deepseekmath}
Zhihong Shao, Peiyi Wang, Qihao Zhu, Runxin Xu, Junxiao Song, Xiao Bi, Haowei Zhang, Mingchuan Zhang, Y.~K. Li, Y. Wu, and Daya Guo.
\newblock {DeepSeekMath}: Pushing the limits of mathematical reasoning in open language models.
\newblock \emph{arXiv preprint arXiv:2402.03300}, 2024.

\bibitem[Sutton and Barto(2018)]{sutton2018reinforcement}
Richard~S. Sutton and Andrew~G. Barto.
\newblock \emph{Reinforcement Learning: An Introduction}.
\newblock MIT Press, Cambridge, MA, 2nd edition, 2018.

\bibitem[Wang et~al.(2025{\natexlab{a}})Wang, Feng, Lai, Xu, Li, Ge, Dehghan, Cao, and Huang]{wang2025streambridge}
Haibo Wang, Bo Feng, Zhengfeng Lai, Mingze Xu, Shiyu Li, Weifeng Ge, Afshin Dehghan, Meng Cao, and Ping Huang.
\newblock {StreamBridge}: Turning your offline video large language model into a proactive streaming assistant.
\newblock In \emph{Advances in Neural Information Processing Systems}, 2025{\natexlab{a}}.

\bibitem[Wang et~al.(2025{\natexlab{b}})Wang, Meng, Wang, Liang, Wei, Zhang, and Zhao]{wang2024mmduet}
Yueqian Wang, Xiaojun Meng, Yuxuan Wang, Jianxin Liang, Jiansheng Wei, Huishuai Zhang, and Dongyan Zhao.
\newblock {VideoLLM} knows when to speak: Enhancing time-sensitive video comprehension with video-text duet interaction format.
\newblock In \emph{Findings of the Association for Computational Linguistics: EMNLP 2025}, pages 6338--6359, Suzhou, China, 2025{\natexlab{b}}. Association for Computational Linguistics.

\bibitem[Wang et~al.(2025{\natexlab{c}})Wang, Meng, Wang, Zhang, and Zhao]{wang2025proactivevideoqa}
Yueqian Wang, Xiaojun Meng, Yifan Wang, Huishuai Zhang, and Dongyan Zhao.
\newblock {ProactiveVideoQA}: A comprehensive benchmark evaluating proactive interactions in video large language models.
\newblock \emph{arXiv preprint arXiv:2507.09313}, 2025{\natexlab{c}}.

\bibitem[Wang et~al.(2026)Wang, Liu, Wang, Xu, Wan, Zhang, and Zhao]{wang2025mmduet2}
Yueqian Wang, Songxiang Liu, Disong Wang, Nuo Xu, Guanglu Wan, Huishuai Zhang, and Dongyan Zhao.
\newblock {MMDuet2}: Enhancing proactive interaction of video {MLLMs} with multi-turn reinforcement learning.
\newblock In \emph{International Conference on Learning Representations (ICLR)}, 2026.

\bibitem[Xia et~al.(2026)Xia, Chen, Zhang, Sun, and Zhou]{xia2026streamo}
Jiaer Xia, Peixian Chen, Mengdan Zhang, Xing Sun, and Kaiyang Zhou.
\newblock Streaming video instruction tuning.
\newblock In \emph{Proceedings of the IEEE/CVF Conference on Computer Vision and Pattern Recognition (CVPR)}, pages 31219--31229, 2026.

\bibitem[Yao et~al.(2025)Yao, Li, Wei, Li, Ren, Liu, Ouyang, Wang, Li, Li, Kong, Liu, Zhang, and Sun]{yao2025timechatonline}
Linli Yao, Yicheng Li, Yuancheng Wei, Lei Li, Shuhuai Ren, Yuanxin Liu, Kun Ouyang, Lean Wang, Shicheng Li, Sida Li, Lingpeng Kong, Qi Liu, Yuanxing Zhang, and Xu Sun.
\newblock {TimeChat-Online}: 80\% visual tokens are naturally redundant in streaming videos.
\newblock In \emph{Proceedings of the 33rd ACM International Conference on Multimedia}, pages 10807--10816. Association for Computing Machinery, 2025.

\bibitem[Yu et~al.(2026{\natexlab{a}})Yu, Li, Si, Zhang, Xu, Wang, Dong, Tuo, Zeng, Feng, Wang, Shi, Hu, Yue, Wang, and Yan]{yu2026dopd}
Xinlei Yu, Gen Li, Qingyi Si, Guibin Zhang, Yuqi Xu, Congcong Wang, Shuai Dong, Kaiwen Tuo, Xiangyu Zeng, Kaituo Feng, Qunzhong Wang, Yang Shi, Xiaobin Hu, Xiangyu Yue, Jiaqi Wang, and Shuicheng Yan.
\newblock {DOPD}: Dual on-policy distillation.
\newblock \emph{arXiv preprint arXiv:2606.30626}, 2026{\natexlab{a}}.

\bibitem[Yu et~al.(2026{\natexlab{b}})Yu, Xu, Chen, Zhang, Lu, Yang, Zhang, Yan, and Hu]{yu2026mact}
Xinlei Yu, Chengming Xu, Zhangquan Chen, Yudong Zhang, Shilin Lu, Cheng Yang, Jiangning Zhang, Shuicheng Yan, and Xiaobin Hu.
\newblock Visual document understanding and reasoning: A multi-agent collaboration framework with agent-wise adaptive test-time scaling.
\newblock In \emph{Proceedings of the IEEE/CVF Conference on Computer Vision and Pattern Recognition (CVPR)}, pages 12300--12311, 2026{\natexlab{b}}.

\bibitem[Yuan et~al.(2025)Yuan, Wang, Sun, Zhang, and Lin]{yuan2025tarsier2}
Liping Yuan, Jiawei Wang, Haomiao Sun, Yuchen Zhang, and Yuan Lin.
\newblock {Tarsier2}: Advancing large vision-language models from detailed video description to comprehensive video understanding.
\newblock \emph{arXiv preprint arXiv:2501.07888}, 2025.

\bibitem[Zeng et~al.(2026)Zeng, Yang, Zhang, Zhu, Li, Si, Yao, Ma, Chen, Chen, Shi, Zhou, Li, Zhang, Qin, Yang, Yu, Ouyang, Shao, Wei, Zhou, Gao, Wang, and Wang]{zeng2026onestreamer}
Xiangyu Zeng, Yuandong Yang, Zhiqiu Zhang, Yuhan Zhu, Xinhao Li, Qingyi Si, Dingyu Yao, Changlian Ma, Haoran Chen, Xinyu Chen, Yansong Shi, Junhao Zhou, Yifei Li, Jun Zhang, Chuanyu Qin, Chenxu Yang, Xinlei Yu, Kun Ouyang, Yuchen Shao, Qianshan Wei, Changhai Zhou, Jun Gao, Jiaqi Wang, and Limin Wang.
\newblock {OneStreamer}: Unifying perception, memory, and proactive response in streaming video interaction.
\newblock \emph{arXiv preprint arXiv:2610.01762}, 2026.

\bibitem[Zhang et~al.(2025{\natexlab{a}})Zhang, Wang, Tang, Liu, Feng, and Jin]{zhang2025flashvstream}
Haoji Zhang, Yiqin Wang, Yansong Tang, Yong Liu, Jiashi Feng, and Xiaojie Jin.
\newblock {Flash-VStream}: Efficient real-time understanding for long video streams.
\newblock In \emph{Proceedings of the IEEE/CVF International Conference on Computer Vision (ICCV)}, pages 21059--21069, 2025{\natexlab{a}}.

\bibitem[Zhang et~al.(2025{\natexlab{b}})Zhang, Wu, Li, Li, Ma, Liu, and Li]{zhang2024llavavideo}
Yuanhan Zhang, Jinming Wu, Wei Li, Bo Li, Zejun Ma, Ziwei Liu, and Chunyuan Li.
\newblock {LLaVA-Video}: Video instruction tuning with synthetic data.
\newblock \emph{Transactions on Machine Learning Research}, 2025{\natexlab{b}}.

\end{thebibliography}

\clearpage
\beginappendix
\setcounter{equation}{0}
\renewcommand{\theequation}{\Alph{section}.\arabic{equation}}

\section{Derivation of Turn-Level Credit Assignment}
\label{sec:suppl:credit}

This section derives the reward decomposition and advantage correction used in \cref{sec:method:credit}.
All group averages below refer to the $G$ aligned rollouts of the same streaming instance.

\subsection{Silence-Completed Prefix Scores}
\label{sec:suppl:credit:prefix}

For each sampled trajectory $i$, \cref{eq:prefix-score} retains its first $t$ actions and replaces every remaining action with $\varnothing$.
The scoring context---initial dialogue, decision timestamps, reply windows, and evaluation horizon---remains unchanged.
This fixed horizon preserves an earlier answer's PAUC contribution over the remainder of its reply window, even when no further answers are generated.
It also keeps the denominators of horizon-dependent penalties unchanged.
Silence completion isolates the contribution of retained replies under the full interaction horizon.

At $t=0$, every rollout is replaced by the same all-silent interaction, so $U_{i,0}=U_0$ for all $i$ in the group.
The shared reference $U_0$ incorporates the initial dialogue and cancels in group-relative credit comparisons.
At the other endpoint, all sampled actions are retained, giving $U_{i,T_i}=R_i$.

\subsection{Accrued Reward and Remaining Return}
\label{sec:suppl:credit:decomposition}

Define $r_{i,t}=U_{i,t}-U_{i,t-1}$ as the change in the complete trajectory score when action $t$ is retained in place of silence, with all later actions still silent.
Summing these differences before decision $t$ gives
\begin{equation}
    \begin{aligned}
        C_{i,t-1}
        &=\sum_{s=1}^{t-1}r_{i,s}\\
        &=\sum_{s=1}^{t-1}(U_{i,s}-U_{i,s-1})\\
        &=U_{i,t-1}-U_0.
    \end{aligned}
    \label{eq:suppl:accrued}
\end{equation}
The intermediate prefix scores cancel, leaving the reward accrued by the retained history relative to the all-silent reference.
Summing from decision $t$ through the last turn similarly gives
\begin{equation}
    \begin{aligned}
        F_{i,t}
        &=\sum_{s=t}^{T_i}r_{i,s}\\
        &=\sum_{s=t}^{T_i}(U_{i,s}-U_{i,s-1})\\
        &=U_{i,T_i}-U_{i,t-1}\\
        &=R_i-U_{i,t-1}.
    \end{aligned}
    \label{eq:suppl:remaining}
\end{equation}
Thus, the quantities used in the main paper obey
\begin{equation}
    \begin{aligned}
        r_{i,t}&=U_{i,t}-U_{i,t-1},\\
        C_{i,t-1}&=U_{i,t-1}-U_0,\\
        F_{i,t}&=R_i-U_{i,t-1},\\
        C_{i,t-1}+F_{i,t}&=R_i-U_0.
    \end{aligned}
    \label{eq:accrued-remaining}
\end{equation}
The last identity partitions the original trajectory score, relative to a shared reference, into accrued reward and remaining return.
This exact partition assigns later decisions their remaining-return credit under the original trajectory scorer.
The reference $U_0$ will cancel in the group-relative comparison below.

The telescoping decomposition is exact for non-additive trajectory scorers, including ratio-based repetition penalties.
In particular, a new reply can change both the numerator and denominator of a ratio-based penalty, such as the repetition fraction $D_i^{\mathrm{rep}}$ in \cref{eq:trajectory-reward}.
Computing each $U_{i,t}$ with the complete scoring function includes both changes in $r_{i,t}$; summing the increments therefore recovers the score difference exactly.
The prefix computation reuses cached quality and repetition assessments because retained replies and their preceding histories remain identical.

\subsection{Group-Relative Credit Correction}
\label{sec:suppl:credit:advantage}

Let $\bar U_{t-1}=G^{-1}\sum_j U_{j,t-1}$.
Because the all-silent reference is shared within the group,
\begin{equation}
    \begin{aligned}
        \bar C_{t-1}&=\bar U_{t-1}-U_0,\\
        \bar F_t&=\bar R-\bar U_{t-1}.
    \end{aligned}
    \label{eq:suppl:group-means}
\end{equation}
Consequently, centering the remaining return gives
\begin{equation}
    \begin{aligned}
        F_{i,t}-\bar F_t
        &=(R_i-U_{i,t-1})-(\bar R-\bar U_{t-1})\\
        &=(R_i-\bar R)-(U_{i,t-1}-\bar U_{t-1})\\
        &=(R_i-\bar R)-(C_{i,t-1}-\bar C_{t-1}).
    \end{aligned}
    \label{eq:suppl:centered-return}
\end{equation}
Dividing by the same terminal-reward scale as GRPO yields
\begin{equation}
    \frac{F_{i,t}-\bar F_t}{\sigma_R+\epsilon}
    =A_i^{\mathrm{global}}
    -\frac{C_{i,t-1}-\bar C_{t-1}}{\sigma_R+\epsilon}.
    \label{eq:suppl:credit-correction}
\end{equation}
Applying the same clipping operator to both sides gives \cref{eq:turnwise-advantage}.
The correction removes above- or below-average reward accrued before the decision, while retaining a common normalization scale across turns.
At the first turn, $C_{i,0}=\bar C_0=0$, so the corrected and global signals coincide before clipping.

TurnWise compares sampled continuation returns $F_{i,t}$ at aligned decision times, using each rollout's own dialogue history and subtracting its accrued-reward reference.

\subsection{Interpolation Strength and Warm-Up}
\label{sec:suppl:credit:alpha}

For $\alpha\in[0,1]$, scaling the accrued-reward correction gives the unclipped signal
\begin{equation}
    \begin{aligned}
        \widetilde A_{i,t}^{(\alpha)}
        &=A_i^{\mathrm{global}}
          -\alpha\frac{C_{i,t-1}-\bar C_{t-1}}{\sigma_R+\epsilon}\\
        &=(1-\alpha)A_i^{\mathrm{global}}
          +\alpha\frac{F_{i,t}-\bar F_t}{\sigma_R+\epsilon}.
    \end{aligned}
    \label{eq:suppl:alpha-equivalence}
\end{equation}
The first line is the scaled correction used in \cref{eq:alpha-advantage}; the second follows directly from \cref{eq:suppl:credit-correction} and gives its equivalent interpolation form.
For example, $\alpha=0.5$ removes half of the accrued-reward correction and equally weights the global and remaining-return signals before clipping.
The interpolation preserves the trajectory scorer and rollout set while controlling the fraction of accrued-reward credit removed.
We interpolate the unclipped signals and then apply the shared clipping operator.

The training schedule gradually introduces this correction:
\begin{equation}
    \alpha(s)=\min\left\{1,\frac{s}{\lceil0.1S\rceil}\right\},
    \label{eq:suppl:alpha-schedule}
\end{equation}
where $s$ counts completed optimizer updates and $S$ is the scheduled total.
Training starts with shared, clipped GRPO credit, reaches full turn-specific correction after the warm-up, and retains $\alpha=1$ thereafter.

\section{Policy Optimization Details}
\label{sec:suppl:policy}
\setcounter{equation}{0}

The GRPO token surrogate in \cref{sec:method:loss} follows the probability-ratio clipping construction of PPO~\cite{schulman2017ppo}, defined in \cref{eq:token-surrogate}, with a dual-clipping safeguard.
For $A\geq0$, $\ell(\rho,A)=\ell_{\mathrm{clip}}(\rho,A)$.
For $A<0$, we use
\begin{equation}
    \ell(\rho,A)
    =\min\bigl\{\ell_{\mathrm{clip}}(\rho,A),-cA\bigr\},
    \label{eq:suppl:dual-clipping}
\end{equation}
where $c>1$ bounds the loss for negative advantages; we set $c=3$.

Turn-mean aggregation is computed over the generated action-token spans of each sampled trajectory.
Observations, user text, prompt headers, and padding are excluded from the loss.
Both textual responses and the generated waiting output \texttt{NO REPLY} are included as decision turns.

\end{document}